\documentclass[pdflatex,sn-mathphys-num]{sn-jnl}

\usepackage[utf8]{inputenc}%
\usepackage[T1]{fontenc}%

\usepackage{graphicx}%
\usepackage{multirow}%
\usepackage{amsmath,amssymb,amsfonts}%
\usepackage{amsthm}%
\usepackage{mathrsfs}%
\usepackage[title]{appendix}%
\usepackage{xcolor}%
\usepackage{textcomp}%
\usepackage{manyfoot}%
\usepackage{booktabs}%
\usepackage{algorithm}%
\usepackage{algorithmicx}%
\usepackage{algpseudocode}%
\usepackage{listings}%
\usepackage{hyperref}
\usepackage{float}
\usepackage{caption}
\usepackage{subcaption}

\usepackage[most]{tcolorbox}
\usepackage{tikz}
\usetikzlibrary{trees}

\usepackage{hyperref}

\usepackage{lineno}

\theoremstyle{thmstyleone}%
\theoremstyle{thmstyletwo}%

\theoremstyle{thmstylethree}%

\begin{document}

\title[Article Title]{A three-Level Framework for LLM-Enhanced eXplainable AI: From technical explanations to natural language}


\author*[1]{\fnm{Marilyn} \sur{Bello}~(ORCID: 0000-0003-4540-2508)}\email{mbgarcia@ugr.es}

\author[2]{\fnm{Rafael} \sur{Bello}~(ORCID: 0000-0001-5567-2638)}\email{rbellop@uclv.edu.cu}

\author[2]{\fnm{Mar\'ia-Matilde} \sur{Garc\'ia}~(ORCID: 0000-0001-6712-6351)}\email{mmgarcia@uclv.edu.cu}

\author[3]{\fnm{Ann} \sur{Now\'e}~(ORCID: 0000-0001-6346-4564)}\email{ann.nowe@vub.ai.ac.be}

\author[1]{\fnm{Iv\'an} \sur{Sevillano-García}~(ORCID: 0000-0002-5029-9106)}\email{isevillano@ugr.es}

\author[1]{\fnm{Francisco} \sur{Herrera}~(ORCID: 0000-0002-7283-312X)}\email{herrera@decsai.ugr.es}

\affil*[1]{\orgdiv{Andalusian Research Institute in Data Science and Computational Intelligence}, \orgname{Universidad de Granada}, \orgaddress{\city{Granada}, \country{Spain}}}

\affil[2]{\orgdiv{Department of Computer Science}, \orgname{Universidad Central “Marta Abreu” de Las Villas}, \orgaddress{\city{Santa Clara}, \country{Cuba}}}

\affil[3]{\orgdiv{Artificial Intelligence Lab}, \orgname{Vrije Universiteit Brussel}, \orgaddress{\city{Brussel}, \country{Belgium}}}


\abstract{The growing application of artificial intelligence in sensitive domains has intensified the demand for systems that are not only accurate but also explainable and trustworthy. Although explainable AI (XAI) methods have proliferated, many do not consider the diverse audiences that interact with AI systems: from developers and domain experts to end-users and society. This paper addresses how trust in AI is influenced by the design and delivery of explanations and proposes a multilevel framework that aligns explanations with the epistemic, contextual, and ethical expectations of different stakeholders. The framework consists of three layers: algorithmic and domain-based, human-centered, and social explainability\textcolor{black}{, with Large Language Models serving as crucial mediators that transform technical outputs of AI explanations into accessible, contextual narratives across all levels. We show how LLMs enable dynamic, conversational explanations that bridge the gap between complex model behavior and human understanding, facilitating interactive dialogue and enhancing societal transparency. Through comprehensive case studies, we show how this LLM-enhanced approach achieves technical fidelity, user engagement, and societal accountability, reframing XAI as a dynamic, trust-building process that leverages natural language capabilities to democratize AI explainability.}
}


\maketitle

\section{Introduction}
\label{sec:1}

Explainable Artificial Intelligence (XAI) has become a foundational area of research in the pursuit of trustworthy AI systems. As machine learning models, particularly black-box systems, are increasingly deployed in sensitive and high-impact domains such as healthcare, finance, and justice, the ability to interpret and understand their behavior is no longer optional, but essential. XAI addresses the critical need for transparency, allowing stakeholders to assess model decisions, detect biases, ensure compliance, and build user trust~\cite{herrera2025reflections}. According to Ali et al.~\cite{ali2023explainable}, while significant progress has been made in developing explainability methods, key challenges remain in aligning these techniques with real-world trust requirements. Their work underscores that achieving trustworthy AI requires explainability approaches that are not only technically sound, but also contextually meaningful and user-centered.

Recent work highlights that XAI must go beyond mere model interpretability and embrace a broader, human-centered vision that incorporates the goals, expectations, and values of diverse stakeholders across the AI lifecycle~\cite{herrera2025reflections}. This evolving perspective reframes XAI as a dynamic socio-technical process, not merely a technical add-on. The definition of XAI given in~\cite{arrieta2020explainable}: \textit{``Given an audience, an Explainable AI is one that produces details or reasons to make its functioning clear or easy to understand''} highlights the importance of considering the target audience when generating explanations.

According to Bodria et al.~\cite{bodria2023benchmarking}, explanations in the context of black-box machine learning models are defined as \emph{``human-interpretable descriptions of the internal logic of a model or the reasons for its predictions.''} These explanations aim to make opaque decision-making processes understandable to stakeholders by highlighting relevant input features, representative examples, or simplified decision rules. Their work emphasizes the importance of tailoring explanation strategies to specific tasks and audiences, considering the trade-offs between fidelity, interpretability, and computational cost.
In~\cite{freiesleben2023dear}, a series of misconceptions and fallacies are identified that, according to their criteria, affect the development and application of XAI. Among the key issues highlighted, they point out that there is no universally suitable explanation method for all cases. Furthermore, according to~\cite{vilone2021classification}, identifying the most effective explanation can be challenging. The variety and diversity of XAI methods make it difficult to select the most appropriate one to explain an intelligent system. Hence, it is beneficial to know about the existence of different XAI methods and to have a guide on how to generate such explanations.

There have been several categorization proposals for XAI methods from several perspectives, each of them complementary and providing a different view of the field.In~\cite{dragoni2022knowledge}, three categories are proposed to classify the different approaches within XAI based on the type of explanation they generate. The first category encompasses methods that generate explanations by identifying the parts of the input that are relevant to the solution produced by the intelligent system, called attribution methods. The second category includes methods that provide declarative explanations, which explicitly and formally expose knowledge. The third category comprises methods that produce explanations in natural language, using representations that are more accessible and understandable to humans.

Therefore, a fundamental challenge persists: most existing approaches prioritize technical transparency without adequately considering the diverse cognitive, contextual, and ethical needs of different users. As AI systems permeate high-stakes and socially sensitive domains, there is an urgent need to move beyond one-size-fits-all explanations. Generic or overly technical explanations risk alienating non-expert users, undermining trust, and failing to meet regulatory or societal expectations. What remains lacking is a structured methodology for tailoring explanations to distinct audiences, ensuring that they are not only technically faithful, but also contextually meaningful, ethically grounded, and socially relevant. This paper addresses this gap by proposing a multilevel audience-aware framework for explainability that aligns explanation strategies with the varying goals, knowledge levels, and trust requirements of stakeholders. To this end, we propose a solution to these issues through a multilevel framework, which can be summarized as:

\begin{enumerate}
  
    \item \textbf{Algorithmic and domain-based level:} responsible for ensuring fidelity between the model and the explanation,
    \item \textbf{Human-centered level:} responsible for integrating human and machine expertise, 
    and
    \item \textbf{Social explainability level:} responsible for making the explanation understandable to non-expert users. 
\end{enumerate}    
     
Figure~\ref{fig:c3} shows how these levels are integrated and interact. Our proposal focuses on explaining the predictions of a black-box model for individual data instances. We begin by generating explanations based on the inference process and domain knowledge, where the first level of the framework is applied. Then, we extend this with interactive explanations, where users actively contribute to shaping or refining the interpretation with the second level. Finally, we add a natural language explanation module to enhance the system’s transparency and foster its social acceptance. This natural language module is particularly relevant in the context of Large Language Models (LLMs), which can generate human-readable explanations that are more accessible to non-expert users. \textcolor{black}{Although LLMs are not explainable, their use is justified and reduced to translating the previously generated algorithmic XAI local explanation into natural language, not to explaining the AI as a whole.}

This multilevel approach reflects how AI systems are actually adopted in real-world contexts, where explanations evolve alongside users' familiarity, system sophistication, and broader social demands. Our proposal combines existing XAI techniques with a timeline-oriented framework that helps practitioners align explanation strategies with the gradual development of trust, understanding, and accountability. To our knowledge, this is the first XAI approach to explicitly treat explanation generation as a dynamic, audience-aware process that increases in semantic depth and social relevance over time.

The main contributions of this work are:
\begin{itemize}
    \item \textbf{Novel multilevel XAI framework}: We propose a three-level explainability framework that systematically addresses different stakeholder needs through algorithmic/domain-based, human-centered, and societal transparency approaches.
    \item \textbf{Audience-aware explanation strategy}: We present a dynamic, timeline-oriented approach that aligns explanation complexity and focus with users' evolving familiarity and trust levels in AI systems.
    \item \textbf{LLM integration methodology}: We show how LLMs can be effectively integrated as interpretation layers while maintaining the integrity of underlying XAI techniques and mitigating hallucination risks.
    \item \textbf{Concrete evaluation mechanisms}: We provide specific frameworks for assessing transparency, trust, and ethical considerations.
    \item \textbf{Practical implementation guidance}: We offer actionable insights for practitioners on how to implement audience-centered explanations that evolve with system deployment and user sophistication.
\end{itemize}

The remainder of this paper is organized as follows. Section~\ref{sec:2} reviews the foundational concepts and motivations behind XAI and explores the distinction between formal explanations and audience-centered interpretations. It emphasizes the importance of tailoring explanations to different stakeholder needs, thus motivating the need for a multilevel approach. Section~\ref{sec:4} presents our proposed multilevel explainability framework, consisting of three levels: algorithmic and domain-based, human-centered, and social explainability. Each level is analyzed in terms of its explanatory focus, its alignment with target audiences, and its contribution to fostering trust. {In Section~\ref{sec:examples}, we illustrate the framework with practical examples that shows how it can be applied to real-world AI systems.}
Finally, Section~\ref{sec:5} offers concluding reflections and outlines promising directions for future research, including practical deployment challenges and the evolving role of LLMs in explainability.

\section{Background on Explainable AI and Audience-Centered Explanations}
\label{sec:2}

To lay the foundation for our proposed multilevel framework, this section provides essential background on the landscape of XAI and the importance of tailoring explanations to diverse user needs. First, in Section~\ref{sec:XAI}, we outline the motivations behind XAI, reviewing both model-driven and post-hoc approaches commonly used to interpret complex machine learning systems. This includes a discussion of how explainability contributes to transparency, accountability, and trust. Then, in Section~\ref{sec:23}, we examine the critical role of audiences in the explanation process. We explore how stakeholders, from developers and domain experts to end-users and society, have distinct interpretability needs, and why audience-centered design is essential to create explanations that are not only technically faithful, but also meaningful and actionable.

\subsection{Explainable AI: Models and Motivations}
\label{sec:XAI}

{As AI systems become more complex and pervasive, the need for transparency and interpretability has become increasingly important. The field of XAI has emerged to address this need, focusing on developing methods and techniques that make the functioning and decisions of complex models more transparent and understandable to humans. Depending on the knowledge representation, we can distinguish two categories: \textit{white-box} models and \textit{black-box} models~\cite{kim2015interactive,samek2017explainable}.} 

\begin{itemize}
  \item \textbf{White-box models:} These are models that can be easily interpreted by humans. Examples of these models are linear regression models, decision trees, association rules, and instance-based models, since the reasoning behind their predictions can be easily understood by humans. Each of these models has a clear and interpretable structure, such as coefficients in linear regression or paths in decision trees, which allows users to trace how input features contribute to the final prediction. 

  \item \textbf{Black-box models:} In contrast, these are models that are difficult to interpret, such as deep neural networks, random forests, or support vector machines. These models are often more complex and involve intricate relationships between input features and predictions, making it challenging for humans to understand the reasoning behind their decisions. Even though we have the mathematical formulation of the black-box model, the relationships between input features and the predictions can make it difficult to understand how the model arrived at a specific decision.
\end{itemize}

Although the distinction between white-box and black-box models may seem clear, recent studies have shown that the boundary between these categories is often blurred. For instance, even models traditionally considered interpretable, such as linear regression or decision trees, can become difficult for humans to understand when they involve a large number of features or highly complex structures. Also, the construction of mixed interpretable and non-interpretable designs, such as neuro-symbolic systems, can lead to models that are partially interpretable. 

The development of methods to extract rules from the knowledge collected in artificial neural networks, as described in~\cite{gupta1999generalized}, represents an important precedent in the direction of explainable artificial intelligence. Later, the term XAI was first introduced in~\cite{van2004explainable} as part of academic research that required a comprehensive understanding of the underlying AI system. It was not until~\cite{arrieta2020explainable} that the field of XAI was formally defined as a dedicated research area focused on developing methods and techniques to make the functioning and decisions of complex models more transparent and understandable to humans and focused on the audience.

Focused on explanations, in~\cite{adadi2018peeking}, four reasons for the development of XAI are defined:
\begin{enumerate}
    \item explanations to justify a decision,
    \item explanations for control situations,
    \item explanations to improve a process, and
    \item explanations to help discover knowledge in an application domain.
\end{enumerate}

Each reason has a different goal, a different target audience, and different expectations. XAI must be developed considering these differences, offering a suitable explanation for each case. According to~\cite{ribera2019can}, an explanation process should be developed by offering multiple explanations, considering the different addressees, and building the explanations cooperatively through a human-machine dialogue. They stated that XAI has the following motivations:

\begin{itemize}
    \item \textbf{To evaluate the intelligent system:} To ensure that the system behaves as expected and to detect possible biases.
    \item \textbf{To improve the system's performance:} To increase security against possible attacks.
    \item \textbf{To learn from the system:} To extract the knowledge discovered by the system.
    \item \textbf{To comply with current legislation:} To respond to legal issues.
    \item \textbf{To help the user accept the technology:} To make them understand how the system works and to avoid rejection due to distrust of the system.
\end{itemize}

This led us to focus attention on the audience that must use and manage the explanations. Biecek and Samek~\cite{biecek2024explain} present a position paper with a very interesting division of the XAI area into two complementary cultures that focus attention on the audience.

\begin{itemize}

  \item Human/value-oriented explanations called BLUE XAI (responsiBle models, Legal issues, trUst in predictions, Ethical issues).
  \item Model/validation-oriented explanations approach, known as RED XAI (Research on data, Explore models, Debug models).
\end{itemize}

{ This categorization focuses on the purpose and audience of the explanations, rather than the technical details of the methods themselves. RED XAI emphasizes the need for researchers and developers to understand and improve AI systems, while BLUE XAI focuses on the ethical and legal implications of AI, aiming to build trust and accountability for all potential end-users of AI systems.}

\begin{figure*}[ht!]
    \centering
    \includegraphics[width=1.00\textwidth]{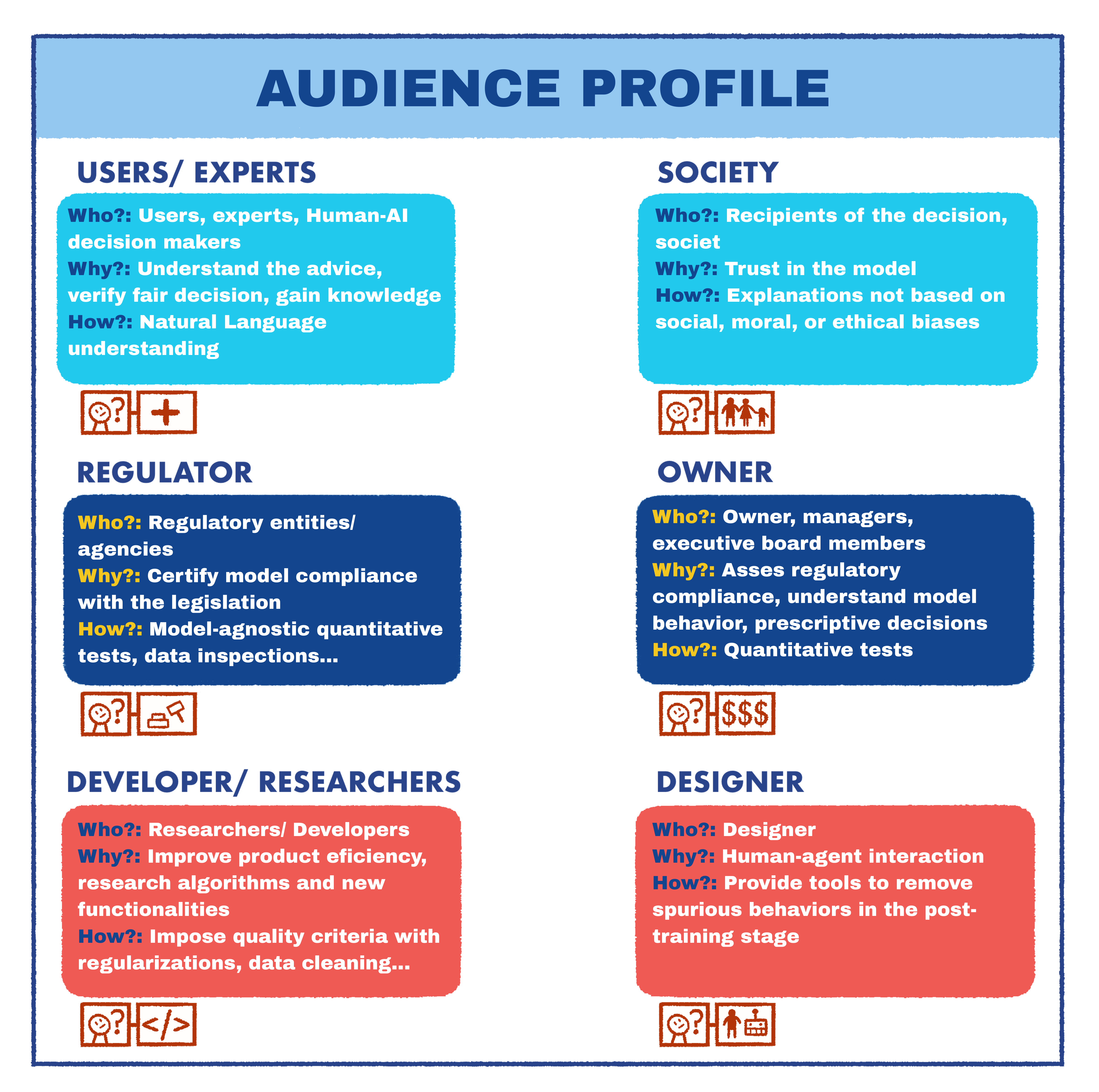}
    \caption{Diagram showing different audience profiles (inspired by figure 4 in~\cite{herrera2025reflections}).}
    \label{fig:audience}
\end{figure*}

As noted by the authors, RED XAI emerges as a compelling but relatively untapped dimension within the broader XAI landscape. This perspective prioritizes in-depth analysis of data, thorough examination of model behavior, and effective debugging to support system dependability. In Figure~\ref{fig:audience}, this is visually represented by the red hue, which corresponds to technical stakeholders such as researchers, developers, and system designers. The figure also includes two shades of blue, symbolizing varying degrees of expertise among audiences focused on human-centric values.

The RED XAI paradigm holds substantial promise for enhancing the robustness and safety of AI technologies, offering a rich area for further investigation. By equipping model developers with targeted feedback, especially during model development and training, RED XAI provides practical insights that can improve system performance and transparency. Its commitment to aligning explanations with both data integrity and the internal mechanics of the model positions it as a valuable instrument for cultivating trust and promoting accountability across AI-driven applications.

In~\cite{haresamudram2023three}, a stakeholder interest map was established. It includes six levels of audience: Developer, Designer, Owner, User, Regulator, Society. The six levels of audience mentioned above are integrated under the question of ``\textit{why this distinction is made?}'' They are shown in Figure~\ref{fig:audience}. For example, for users, the distinction is made in ``\textit{understand the advice, verify fair decisions, and gain scientific knowledge.}''

Table~\ref{table:2} shows the different target audiences identified in the literature, along with examples of each audience and references to the works where they are mentioned. In this table, we fuse the two first categories associated with RED XAI.

\begin{table}[!ht]
\centering
\caption{Different target audiences identified in the literature.}
    \label{table:2}
{\small
\begin{tabular}{|p{2.5cm}|p{2.5cm}|p{2.5cm}|p{3.5cm}|}
   \hline
        Target audience & Examples & Benefits &Reference  \\
        \hline
        developers, researcher \& designers & designers, data scientists, AI researchers, etc. & Bias detection and correction &~\cite{montavon2018methods},~\cite{preece2018stakeholders},~\cite{mittelstadt2019explaining},~\cite{ribera2019can}, ~\cite{zhou2019towards},~\cite{arrieta2020explainable},~\cite{langer2021we},~\cite{bellucci2022combining}, ~\cite{meske2022explainable} \\
        \hline
        owners             & managers, executives, CEOs, etc. & Direct knowledge about how its business works &~\cite{arrieta2020explainable}, ~\cite{langer2021we},~\cite{meske2022explainable} \\
        \hline
        regulators    & politicians, lawyers, ethicists, etc. & Ensuring that a certain ethical or moral bias or criterion is used, or not used, in a particular AI tool &~\cite{preece2018stakeholders}, ~\cite{miller2019explanation},~\cite{arrieta2020explainable},~\cite{floridi2021ethical},~\cite{langer2021we} \\                  
        \hline
        domain experts       & doctors, bankers, etc.                             & Ensure that critical decisions are based on well-established criteria & ~\cite{gilpin2018explaining},~\cite{preece2018stakeholders},~\cite{ribera2019can}, ~\cite{arrieta2020explainable},~\cite{langer2021we}, ~\cite{bellucci2022combining},~\cite{meske2022explainable} \\
        \hline
        end-users and society    & lay users, affected parties, patients, bank customers, etc. & Understanding, trust, fairness &~\cite{lepri2018fair},~\cite{lipton2018mythos}, ~\cite{preece2018stakeholders},~\cite{ribera2019can},~\cite{arrieta2020explainable}, ~\cite{langer2021we},~\cite{bellucci2022combining},~\cite{meske2022explainable} \\
        \hline
        \end{tabular}
}%
\end{table}

A reasonable explanation to a system developer may not be a good explanation to a user or domain expert. Although there is no absolute truth about the number and types of stakeholders, it is clear that they exist and their nature significantly influences the explanations required.

As explained in describing the differences between RED XAI and BLUE XAI, a clear distinction must be made between an explanation provided to a researcher and one provided to an end-user. Researchers often develop explanations from their point of view rather than from the point of view of other accreditees (target audiences or stakeholders)~\cite{schoonderwoerd2021human, kim2023should}. 

\subsection{Tailoring Explanations to Stakeholders} 
\label{sec:23}

An interesting taxonomy for organizing XAI methods is presented in~\cite{bodria2023benchmarking}, where the methods are organized in terms of \textit{ input data from the model} (tabular data, image data, text data, time series and graphs) and \textit{output representation of the explanation} (Feature importance~\cite{ribeiro2016should,lundberg2017unified}, Rule-based explanation~\cite{guidotti2018local,guidotti2019factual}, Prototype-based explanations~\cite{kim2016examples,tan2020tree}, and Counterfactual-based explanations~\cite{wachter2017counterfactual,guidotti2022counterfactual}). Some of these methods are based on creating additional models that approximate the original model space and manipulate parts of the input to generate explanations~\cite{ribeiro2016should}. Others are based on quantifying the contribution of the input features to the predicted output using propagation rules~\cite{bach2015pixel} or calculating the prediction gradient to the input features~\cite{selvaraju2017grad}. Alternatively, another approach explains the decisions of complex deep learning architectures based on local reducts of an image~\cite{bello2024reprot}. Each of these approaches has its strengths and weaknesses, and the choice of method depends on the specific context and requirements of the explanation task.

Another criterion used to group XAI methods is according to the information used to generate the explanation.

\begin{itemize}
    \item \textit{Model-based explanations}: This type of explanation has received different denominations: algorithm-centric view~\cite{liao2020questioning}, zero-order explanations~\cite{dazeley2021levels}, model-based explanations~\cite{jiang2022needs}, and static explanations~\cite{lombrozo2006structure}. Some of the most commonly used XAI methods to generate these kinds of explanations are: Local Interpretable Model-Agnostic Explanations (LIME)~\cite{ribeiro2016should}, Anchors~\cite{ribeiro2018anchors}, Local Rule-Based Explanations (LORE)~\cite{guidotti2018local}, Shapley Additive exPlanations (SHAP)~\cite{lundberg2017unified}, Layer-wise Relevance Propagation (LRP)~\cite{bach2015pixel}, Deep Learning Important FeaTures (DeepLIFT)~\cite{shrikumar2017learning}, Gradient-weighted Class Activation Mapping (Grad-CAM)~\cite{selvaraju2017grad} and its variants, SmoothGrad~\cite{smilkov2017smoothgrad}, and Integrated Gradients~\cite{sundararajan2017axiomatic}.
    \item \textit{Domain-based explanations}: Several authors refer to this type of explanation as human-centered explanations~\cite{ehsan2020human,liao2020questioning,schoonderwoerd2021human}. The use of domain knowledge to increase the interpretability of intelligent systems has been studied in different works~\cite{kocielnik2019will,wolf2019explainability,confalonieri2019ontology,tiddi2022knowledge,confalonieri2023multiple}.
    \item \textit{User-based explanations}: This type of explanation can be approached as a Human-Machine Interaction (HCI) process~\cite{amershi2019guidelines,liao2020questioning,costa2022core}. In~\cite{chromik2021human}, the principles for user interfaces that support explanations, called Explanation User Interfaces (XUI), are reviewed. These XUIs allow us to take the static explanations and build understandable explanations based on different users. In~\cite{shneiderman2020bridging}, two types of XUI are described: explanatory and exploratory. The former aims at conveying a single explanation, visually or textually. The latter allows users to explore the behavior of the learned model freely, which is very effective when users can change or influence the model input.  
\end{itemize}

{As we show, there are several XAI methods that can be used to generate explanations. However, the choice of a specific method depends on the specific context and requirements of the explanation task. There is no universal method that can be applied in all cases. Therefore, there is a gap that can be filled with a multilevel approach that considers the different target audiences and the different types of explanation that can be generated.}

In the following, different definitions of explanations are described, showing that there is no consensus on what an explanation is. Since the idea of what the field of XAI should be is incomplete, there is no clear conception of what an explanation actually is. In~\cite{lipton2018mythos}, it is stated that: \textit{``Explanations are ill-defined, finding a consensus on what an explanation is still challenging''}. Some authors in the literature define it as follows.

\begin{itemize}
    \item Textual or visual artifacts that provide a qualitative understanding of the relationship between the components of the instance and the prediction of the model~\cite{ribeiro2016should}.
    \item A collection of interpretable domain features that contribute to a model response for a particular instance~\cite{montavon2018methods}. It is based on the principle that the most relevant input features are those to which the output is most sensitive.
    \item An interaction between two parties: the one that generates the explanation and the addressee~\cite{mucha2021interfaces}.
\end{itemize} 

The explanations are usually oriented to answer different questions~\cite{liao2020questioning}: 
\begin{itemize}
  \renewcommand\labelitemi{-}

  \item How is the solution obtained?

  \item Why is that the inference made by the intelligent system? 

  \item Why is another given answer not obtained? how would the system respond to a given change? 

  \item What minimum changes must be made to an input to obtain another prediction? 

  \item How far can an input be modified to keep the exact prediction?
\end{itemize}

{Each of these questions requires a different approach, even when some of them can be answered with the same method.}

Explanations can be presented in several ways. The most common ones, according to several authors~\cite{arrieta2020explainable,vilone2021classification}, are:

\begin{enumerate}
    \item Numerical explanations, which measure the contribution of the input variables to the model output. This contribution quantifies the sensitivity of a feature to the model output by ranking the variables that most influenced the solution. 
    \item Textual explanations, which use natural language to ensure that humans can easily understand the explanation. 
    \item Visual explanations, which uses visual artifacts that highlight specific areas of an image or words in a text.
    \item Explanations by example, which extracts prototypical samples on data previously validated that relate to the output generated by a given model.
    \item Counterfactual examples, which presents cases that are similar to the original input but lead to a different output. This type of explanation is useful for understanding the boundaries of the model's decision-making process.
    \item Explanations are simplified, which builds an interpretable model on top of the black-box model. This category of methods aims to replace the decision function with an approximate self-explanatory local model.      
    \item Rules, which illustrates the decision-making process of a model step by step. They are schematic and logical, more structured than visual or textual explanations.
\end{enumerate}

The analysis presented in~\cite{kim2023should} suggests that the target audience that does not have professional knowledge of technology prefers explanations based on a combined textual and visual representation. Regardless of the form of explanation representation, we can distinguish two ways of presenting explanations: statically, where explanations do not change in response to any user demand, and interactively, where different types of explanation are iteratively considered with the aim of satisfying different user demands~\cite{arya2019one}. Those explanation presentations are not mutually exclusive, and they can be combined to provide a more comprehensive understanding of the model's behavior.

\textcolor{black}{In summary, the literature shows that there is no consensus on what an explanation is, how it should be generated, and how it should be presented. However, there exist several proposals to take into account before offering an explanation. In~\cite{tchuente2024methodological}, a list of 6 qualitative needs are identified on which an explanation must be based. This list includes a need of human explanation on why our specific task needs AI to be fullfilled, a need of human explanation of the performance metrics or the need of validate the XAI explanation. In~\cite{10264877}, the difference between faithfulness and plausability is discussed. To accomplish the discussion, the authors differentiate between the \textit{evidence} provided by the model to generate the explanation, the \textit{interpretation} of the explanation generated from the evidence and the \textit{faithfulness} of the explanation compared to the real behavior of the model. These approximations are complementary and can be used to guide the development of XAI implementations in real cases. }

\section{A Multilevel Framework for Audience-Aware Explainability}
\label{sec:4}

{This section presents the multilevel explainability framework that generates explanations that are relevant to the target audience and the type of explanation to be generated. This framework is composed of three explainability components: (1) algorithmic and domain-based, which is responsible for generating explanations based on the learned model and domain knowledge, (2) human-centered, which can be seen as an interactive process that integrates human and machine expertise, and (3) societal centered, designed to generate explanations that are tailored to the needs and expectations of different stakeholders. These components structure the explanation process for an intelligent system, from model behavior to user understanding. The central argument of this approach is that explainability is not merely a technical feature, but a relational and ethical requirement for building trust in AI. A layered approach to explanation, rather than a single method, is necessary for the adoption of trustworthy AI}.
Ilustrative examples are introduced to show the potential usefulness of the multilevel framework.

\subsection{General description of the Multilevel Framework for Explainability}

{In this section, we show how the three-level framework to generate explanations works and how they interact. Each level is designed to address different aspects of explainability. Those levels are defined as follows:}

\begin{itemize}
    \item The first level, \textit{Algorithmic and domain-informed explainability}, is responsible for ensuring that the explanation is built using information from the learned model and including domain knowledge.
    
    This level focuses on generating explanations that are faithful to the model's behavior and aligned with domain-specific knowledge. It uses techniques such as feature importance, rule-based explanations, and example-based explanations to provide a clear understanding of how the model makes decisions. This perspective aligns closely with the goals of RED XAI, which emphasizes the importance of deeply understanding model internals, validating outputs through data exploration, and enabling systematic debugging. RED XAI supports this level by guiding developers and researchers in generating technically sound and diagnostically useful explanations, especially during the development and training phases. Its emphasis on interpretability grounded in algorithmic and data-level evidence enhances the reliability of the explanations produced, making this level not only the foundation of the framework but also a key contributor to AI robustness and trustworthiness.

    \item In the second level, \textit{Human-centered and interactive explainability}, the explanation is built through the interaction between the intelligent system and stakeholders. This level emphasizes the importance of user-centered design, where explanations are tailored to the specific needs and expectations of different stakeholders. This level has the goal of integrating human and machine expertise, allowing users to refine explanations based on their knowledge and experience.  This level is intended to also allow expert users to refine the behavior of the system, providing feedback to the system and allowing it.

    This human-centered approach resonates with the principles of BLUE XAI, which focuses on aligning explanations with human values, ethical considerations, and the need to build trust among stakeholders by making AI systems understandable, fair, and accountable.

    \item The third level, \textit{Societal transparency and trust}, aims to reach a level of transparency that allows evaluation of the ethical and legal aspects at the level of end-users and society. This level has the objective of ensuring that the explanations are understandable to non-expert users and making sure that the system's behavior is aligned with societal values and norms. It uses natural language explanations to communicate the model's behavior based on the explanations generated in the first two levels. Moreover, it also fully embodies the BLUE XAI philosophy by prioritizing transparency, ethical alignment, and trust for non-expert users and society at large, ensuring that AI systems are not only explainable but also socially accountable.

\end{itemize}

The relation between the three levels is shown in Figure~\ref{fig:c3}. The first level provides the foundation for the explanations, which are then refined and adapted at the second level through stakeholder interaction. This interaction may involve feedback loops, where stakeholders can ask questions or request clarifications, leading to more tailored and relevant explanations. The third level builds on the explanations generated in the first two levels, ensuring that they are socially acceptable and aligned with ethical standards. This level emphasizes the importance of the societal context and the need for explanations that resonate with a broader audience, including end users and society.

\textcolor{black}{Our approach considers that users will engage with levels 2 and 3 as needed until they achieve satisfactory understanding. If users remain unsatisfied after initial explanations, they can continue to request clarifications and refinements through the interactive components of the framework. However, if after comprehensive engagement across all three levels users are still not satisfied, the system requires renewed human supervision and potential restructuring of the explanation strategy. This feedback mechanism ensures that the framework remains responsive to evolving user needs and maintains its effectiveness across diverse stakeholder groups.}

\textcolor{black}{The use of the three proposed levels is essential and complementary. 
Level 1 trust is ensured by the faithfulness explanations provided by the XAI algorithm and the model behavior aligned with domain-specific knowledge.
Level 2 proposes a feedback loop to align AI with human criteria.
Level 3 brings explanations closer to people who are not familiar with XAI explanations. The combined use of all three ensures trust through Level 1 algorithms, alignment through Level 2 feedback, and understandability by feeding explanations into an LLM that dialogs with the user through Level 3.}

This multilevel approach reflects how intelligent systems are actually adopted in real world contexts, where explanations evolve alongside users' familiarity, system sophistication, and broader social demands. Our proposal combines existing XAI techniques with a timeline-oriented framework that helps practitioners align explanation strategies with the gradual development of trust, understanding, and accountability.
\textcolor{black}{In contrast to other framework proposals, that focus on several aspects that an AI system should fulfill to be explainable, our proposal emphasizes the importance of the interaction between the system and its users throughout the explanation process.}
To our knowledge, this is the first XAI approach to explicitly treat explanation generation as a dynamic, audience-aware process that increases in semantic depth and social relevance over time.
The following subsections discuss the three explainability levels of the proposed framework.

\subsection{Level 1: Algorithmic and Domain-Informed Explainability}
\label{sec:stage1}

\begin{figure}[!ht]
    \centering
    \includegraphics[width=\textwidth]{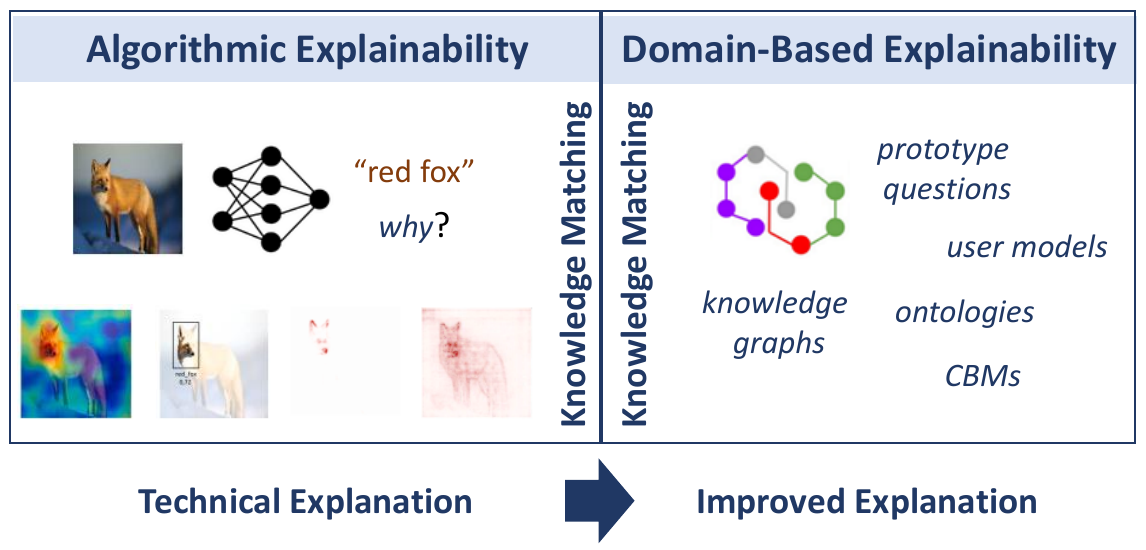}
    \caption{Algorithmic and Domain-Based Explainability level.}
    \label{fig:c1}
\end{figure}

The first level of our framework focuses on generating explanations based directly on the trained model and incorporating knowledge of the application domain. This level is essential for ensuring that the explanations are faithful to the model's behavior and aligned with domain-specific knowledge. We show in Figure~\ref{fig:c1} an illustration of this level, which highlights the relationship between the model, the domain knowledge, and the generated explanations. 

Three main families of methods are used to generate explanations based directly on the trained model:

\begin{itemize}
    \item \textbf{Inherently interpretable models:} These include decision trees, linear regression, rule-based systems, and case-based reasoning. They offer direct transparency into how decisions are made, although complexity can still reduce interpretability.
    \item \textbf{Post-hoc explainability:} These methods approximate the reasoning of black-box models. They can be further divided into:
      \begin{itemize}
        \item \textbf{Feature attribution methods:} Examples include SHAP~\cite{lundberg2017unified}, LIME~\cite{ribeiro2016should}, LRP~\cite{bach2015pixel}, and Grad-CAM~\cite{selvaraju2017grad}, which attribute influence to input features on model output. These approaches can reveal feature importance, generate heatmaps, or highlight relevant text.
        \item \textbf{Example-based explanations:} These methods provide explanations by presenting specific instances or examples that illustrate the model's behavior, such as prototypes or counterfactuals. They help users understand the decision-making process by showing similar cases or hypothetical changes that lead to different outcomes.
      \end{itemize}
\end{itemize}

One way to improve these algorithmic explanations is to supplement them with domain knowledge~\cite{naiseh2023different}. Domain knowledge enriches the explanations with semantic information by introducing vocabulary and terminology from the application domain and verifying the consistency of the explanations with the expectations and experiences of the domain experts.

To generate these explanations, a wide variety of open-source toolkits can be used to support this level. In Table~\ref{tab:stage1-toolkits}, we summarize a selection of the most prominent ones. The choice of toolkit depends on the specific requirements of the task, such as the type of data (tabular, text, image, etc.), the model architecture (neural networks, random forests, etc.), and the desired explanation format (local, global, counterfactual, etc.). These toolkits enhance knowledge by understanding the application domain and the expectations of the target audience.

\begin{table}[!ht]
\centering
\caption{Popular XAI toolkits for algorithmic explainability.}
\label{tab:stage1-toolkits}
{\small
\begin{tabular}{|p{2cm}|p{3.5cm}|p{5.5cm}|}
   \hline
   \textbf{Toolkit} & \textbf{Main Techniques} & \textbf{Key Features} \\
   \hline
   \texttt{AIX360} & Counterfactuals, ProtoDash, LIME, SHAP & Multi-modal support (tabular, text, image, time series); Faithfulness and Monotonicity evaluation \\
   \hline
   \texttt{Alibi} & Anchors, SHAP, Integrated Gradients & Model-agnostic and model-specific; local and global explanations; Multi-modal support (tabular, text, image); Trust Score and Linearity evaluation \\
   \hline
   \texttt{Captum} & Grad-CAM, Integrated Gradient, DeepLift, SHAP & PyTorch-native; interactive visualization tool; Multi-modal support (image, text, audio, video); Scalability, Infidelity, and Sensitivity evaluation \\
   \hline
   \texttt{iNNvestigate} & Saliency, GuidedBackprop, SmoothGrad, IntegratedGradients, PatternAttribution, and LRP & Perturbation analysis \\
   \hline
   \texttt{InterpretML} & GAMs, linear models, rules, SHAP, LIME & Unified API for white-box and black-box; Explainable Boosting Machine \\
   \hline
   \texttt{OmniXAI} & Grad-CAM, Grad-CAM++, Score-CAM, Layer-CAM & Model-agnostic and model-specific; Multi-modal support (tabular, images, texts, and time series); scikitlearn; PyTorch/TensorFlow \\
   \hline
   \texttt{DALEX} & SHAP, LIME, Ceteris Paribus & Model-agnostic; Tabular support; Python and R; Interactive visualizations \\
   \hline
   \texttt{Xplique} & Attribution methods, Feature Visualization, Example-based & Model-agnostic and model-specific; Multi-modal support (tabular, text, image); Concept-based explanations \\
   \hline
\end{tabular}
}%
\end{table}

Among the alternatives to include domain knowledge in the explanation generation process, there are several approaches that should be considered.

\begin{itemize}
    \item \textbf{User models:} This approach involves building frameworks that integrate multiple user interfaces and generate multimodal explanations, facilitating communication and understanding for different types of users. For example, ExAID~\cite{lucieri2022exaid} provides such a framework to support explainability in medical imaging.
    \item \textbf{Prototype questions:} In this alternative, the system maintains a repository of typical questions that users might ask. Explanations are then generated as answers to anticipated questions, aligning the explanation process with user expectations and needs.
    \item \textbf{High-level concepts:} This approach incorporates domain knowledge into the explanation process through ontologies and knowledge graphs, enabling explanations that use terminology and relationships familiar to domain experts. Examples include the use of knowledge graphs in XAI~\cite{lecue2020role,rajabi2022knowledge,tiddi2022knowledge} and ontology-based explanations~\cite{bellucci2022combining}. XAI platforms such as Doctor XAI~\cite{panigutti2020doctor}, Thales XAI~\cite{lecue2019thales}, and TREPAN Reloaded~\cite{confalonieri2019trepan} support this type of explanation.
    \item \textbf{Concept Bottleneck Models~(CBMs) and concept-based explanations:} This line of work structures the learning process by introducing an explicit layer of human-interpretable concepts between the input and the prediction~\cite{yang2023language,srivastava2024vlg}. Techniques such as Concept Relevance Propagation (CRP)~\cite{achtibat2023attribution} translate traditional attribution maps into concept-based explanations, allowing the relevance of a prediction to be traced back to predefined concepts. Prototypical Concept-based Explanations (PCX)~\cite{dreyer2024understanding} further extend this idea by automatically identifying latent concepts and illustrating them with representative examples from the data, thus providing more intuitive and semantically meaningful explanations.
\end{itemize}

All of these alternatives can provide explanations that align with the behavior of the learned model and can be complemented by each other to generate more comprehensive explanations. The choice of the approach depends on the specific context and requirements of the explanation task, \textcolor{black}{as well as using all of them in a complementary way or decision tree interpretation, and the knowledge domain.}

Although these explanations confer algorithmic explanations and are based on domain knowledge, there are still some limitations that need to be addressed, such as the need for explanations that are tailored to the specific needs and expectations of different stakeholders. This is where the second level of our framework comes into play, focusing on human-centered explainability.

\textcolor{black}{It is crucial to emphasize that in our framework, the limitations in explanation generation are primarily associated with the algorithmic level (Level 1), not with the LLMs used in subsequent levels. The fidelity, accuracy, and technical soundness of explanations depend on the quality of the XAI methods employed at this foundational level—such as SHAP, LIME, or decision tree interpretations. LLMs serve as translators that convert these algorithmically generated explanations into human-readable language, but they do not generate the core explanatory content. This architectural separation ensures that the technical integrity of explanations is preserved while enhancing their accessibility.}

\subsection{Level 2: Human-Centered and Interactive Explainability}
\label{sec:stage3}

\begin{figure}[!ht]
    \centering
    \includegraphics[width=\textwidth]{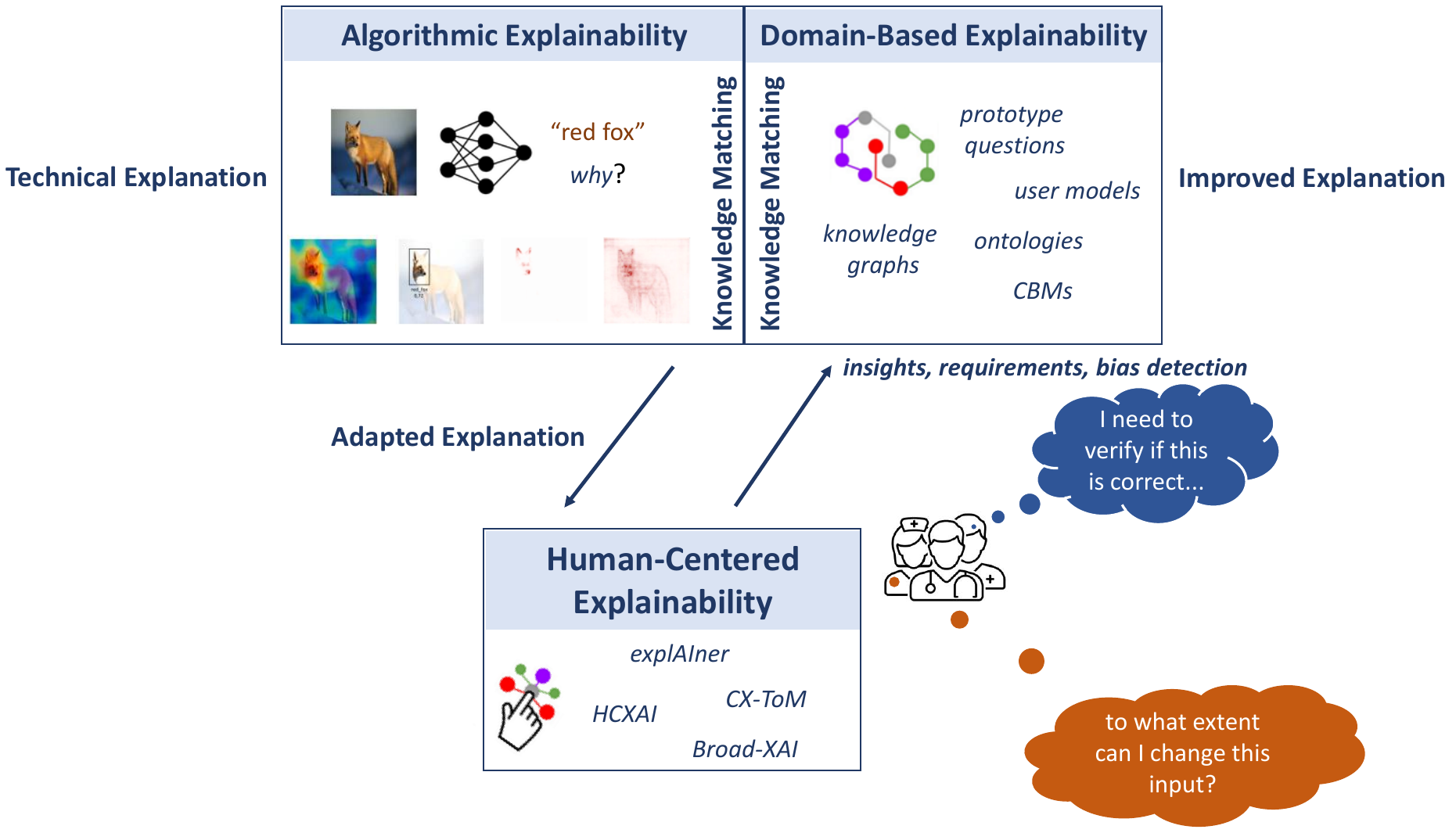}
    \caption{Human-Centered Explainability level.}
    \label{fig:c2}
\end{figure}

In this section, we describe the second level of our framework. This level focuses on generating explanations through interaction with stakeholders. It recognizes that explanations are dynamic processes that evolve through dialogue between the intelligent system and its users. This dialogue allows the system to eventually reach a human-feedback loop, where the system can adapt its explanations based on user feedback and preferences. In Figure~\ref{fig:c2}, we illustrate the interaction between the intelligent system and stakeholders, highlighting the importance of user feedback in shaping the explanations.

The generation of explanations at this level can be developed as an interactive process structured around domain analysis, requirements analysis, and interaction design, where it is possible to establish a bidirectional relationship with the first level to reconstruct the explanation interactively according to the user's point of view and interests. However, according to~\cite{riveiro2021s}, designing an interactive explanation system that satisfies the interests of different target audiences is a complex task, but it can be gradually and partially developed in such a way that it meets the expectations of stakeholders.

This can be done following two approaches: one in which explanation patterns are derived from use cases, and another in which the explanation of each case is built interactively. In the first case, it is possible to get a preliminary insight into the kind of explanation that users expect to receive from the system by asking users to provide explanations themselves based on the use cases. In the other case, interactive interfaces allow users to freely explore the algorithms by changing the values of the model parameters and showing the impact of these changes on the inferences made by the intelligent system. In addition, users can obtain information by analyzing different combinations of inputs. The interaction in this stage is focused on experts or, at least, on users with some knowledge of the system, as managing the interaction requires a certain level of understanding. Some results that support this stage and that we recommend consulting for its development are as follows.

\begin{itemize}
    \item In~\cite{spinner2019explainer}, the \textit{explAInerframework } is presented, which integrates visual and interactive capabilities to facilitate users' understanding of the learned models, diagnosis of their limitations, refinement, and optimization.
    \item In~\cite{ehsan2020human}, the \textit{Human-Centered Explainable AI} approach is presented, aimed at generating explanations for nontechnical end-users. 
    \item In~\cite{dazeley2021levels}, the \textit{Broad Explainable AI}  approach is presented, in which the construction of explanations is part of a conversational explanation system interactive with humans.
    \item In~\cite{akula2022cx}, the \textit{Counterfactual eXplanations with Theory-of-Mind} framework is presented, which generates a sequence of explanations through a human-machine dialogue.
\end{itemize}

These approaches allow expert users to refine the behavior of the system, providing feedback to it.
\textcolor{black}{Additionally, this level serves as a critical checkpoint for risk control, where experts can identify and prevent harmful or misleading explanations through human-machine collaboration, ensuring that the model's behavior aligns with proper ethical and domain-specific standards.}
Since these XAI systems are designed to help end users decide certain actions, these explanations should at least be understandable to users with domain knowledge, but not necessarily with technical knowledge. In this context, the explanations generated at this level are more user-centered and tailored to the specific needs and expectations of certain stakeholders. These cases may be one of the critical societal domains such as healthcare, finance, education, and public administration. Each domain expert may use the XAI system to decide on tasks in which far-reaching consequences may affect individuals and communities.

However, even if the expert understands the model, or the explanation provided by the system, it is not guaranteed that the social impact of the decision is understood by non-expert audiences. In this context, explanations must go beyond technical transparency or user-level understanding. They must address broader societal concerns such as fairness, accountability, trust, and ethical alignment. For this purpose, the third level of the framework is designed to extend the explanations to nonexpert users and society, ensuring that the explanations are accessible and meaningful for all users.

\textcolor{black}{When implementing this framework in high-risk domains such as healthcare and finance, where sensitive data may be involved, it is crucial to consider privacy protection requirements. In such critical applications, we strongly recommend using on-premise LLMs instead of online services to mitigate privacy concerns and ensure compliance with established ethical standards and data protection regulations.}

\textcolor{black}{Another important consideration when incorporating LLMs into explainability frameworks is the potential risk of hallucinations—instances where the model generates plausible but factually incorrect information~\cite{huang2025survey}. However, our multilevel approach significantly mitigates this risk compared to frameworks that rely exclusively on LLMs for explanation generation. In our framework, the original algorithmic explanation is provided by Level 1 through established XAI methods, ensuring a solid foundation of technical accuracy. The LLM operates by interpreting and translating this pre-existing explanation rather than creating explanatory content from scratch. Additionally, the prompts sent to LLMs contain enriched contextual information including domain knowledge, ethical constraints, user profiles, and regulatory considerations, enabling more appropriate and tailored explanations. This design reduces the likelihood of hallucinations since the LLM is constrained by the algorithmic explanation as its source material. Furthermore, any potential hallucinations become more readily detectable when users can compare the natural language explanations with the underlying algorithmic explanations, providing an additional layer of verification that would be absent if explanations were generated directly by LLMs without algorithmic grounding.}

\subsection{Level 3: Understandability and dialogue for Societal Trust}
\label{sec:stage4}

In this section, we present the final level of our XAI framework. This level aims to bridge the gap between AI outputs and collective human values by fostering public oversight, enabling informed debate, and supporting democratic legitimacy. They are essential for cultivating trust among diverse end users, ensuring that AI systems operate in socially acceptable, legally compliant, and culturally sensitive ways.

According to~\cite{haresamudram2023three}, ``\textit{AI systems are not just algorithms but, through their use, give rise to complex interactions between individuals and devices within specific contexts and environments that, in turn, are governed by social norms, cultural expectations, and laws}''. It includes six levels of audience: Developer, Designer, Regulator, Owner, User, and Society~\cite{herrera2025reflections}. The six audience levels mentioned above are integrated under the question: Why is this distinction made?

{In this light, explanations must evolve from merely technical or interactive narratives to those that address societal-level impacts. This third level acknowledges that the widespread adoption of AI technologies demands a layer of explanation that speaks to social legitimacy, ethical acceptability, and cultural resonance. For example, for users, the distinction is made in \textit{``understand the advice, verify fair decisions, and gain scientific knowledge''}. For society, the distinction is made in \textit{``trust in the model itself, rely on the decision''}.}

Societal explanations for users and society are fundamental. We can consider the following aspects: 

\begin{itemize}
    \item \textbf{Societal explanation as collective accountability:} Although individual explanations may suffice for personal decision making, societal explanations function as tools for public oversight. Mechanisms such as citizen juries, regulatory sandboxes, and participatory audits are emerging as essential tools in this space. They embed the explanation in institutional practices, allowing affected communities to contest, reinterpret, or reshape the role of AI in their lives. This is related to \textit{understanding the advice, verifying fair decisions, and gaining scientific knowledge}.
    \item \textbf{Reconceptualizing explanation as social dialogue:} Social explanations must inform and engage. They need to support civic trust and encourage critical discourse around AI. This is related to \textit{trusting the model itself and relying on the decision}.
\end{itemize}

\begin{figure}[!ht]
    \centering
    \includegraphics[width=\textwidth]{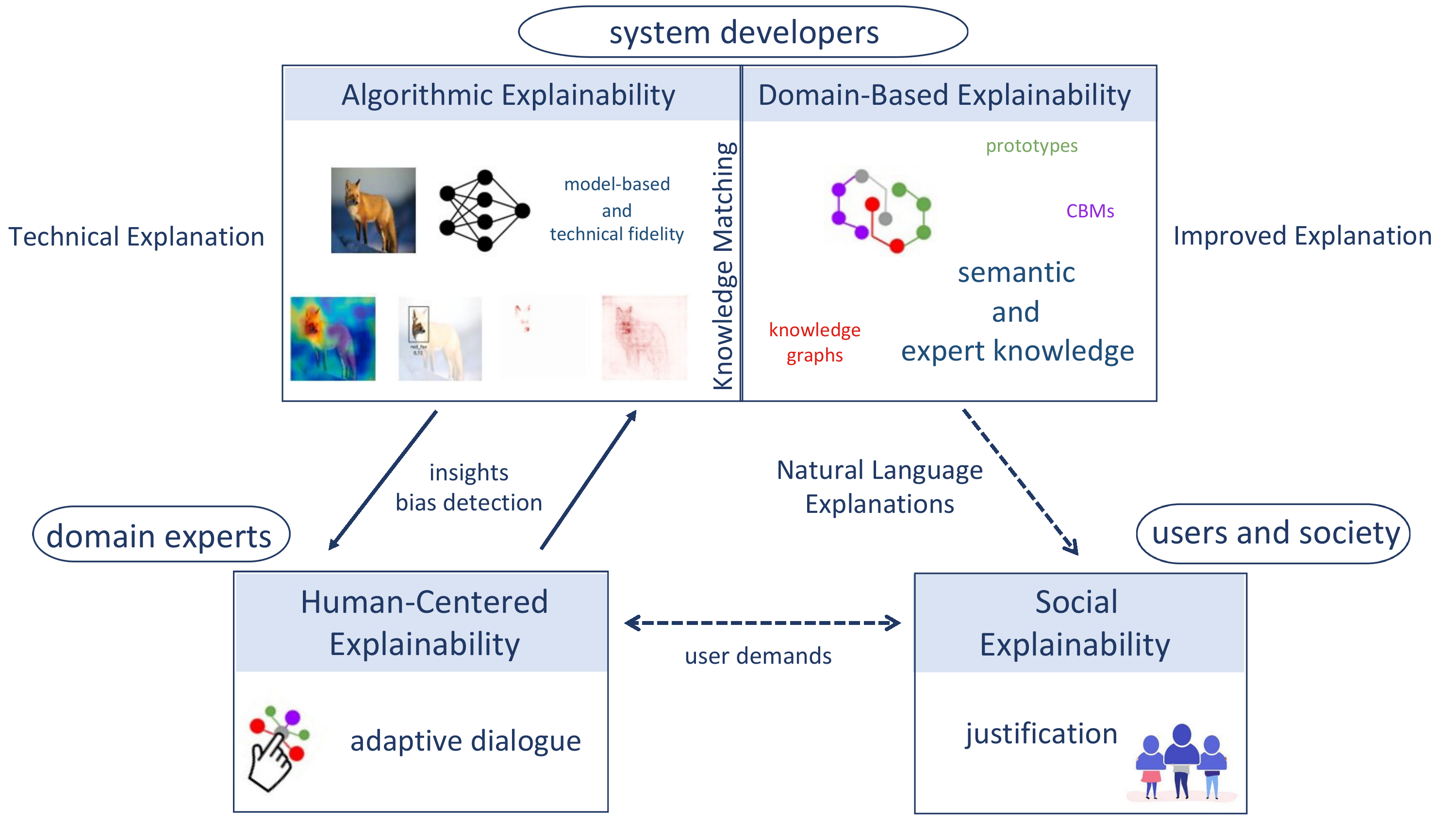}
    \caption{Understandability and dialogue level.}
    \label{fig:c3}
\end{figure}

In essence, societal explanations represent the final and broadest purpose of XAI, as we show in Figure~\ref{fig:c3}. They demand that developers, policymakers, and civil society collaborate to communicate what AI systems do and to shape what they should do in a democratic society. They serve not just as an epistemic function, but as a moral and civic one: supporting inclusive, reflective, and ethically grounded AI adoption.

\textcolor{black}{The urgency of societal-level explainability has intensified with the advent of global AI regulations, such as the EU AI Act (2024)~\cite{eu-269-2014}, which establish transparency and accountability as legal imperatives crucial for public trust and democratic governance. Our framework’s third level addresses these requirements by enabling accountability mechanisms that extend beyond technical compliance—supporting independent audits, regulatory reporting, and citizen participation—while aligning with the EU AI Act’s emphasis on human oversight and democratic control, ultimately transforming technical AI outputs into socially meaningful narratives that inform public discourse and policymaking.}

\textcolor{black}{It is crucial to clarify how trust and ethical considerations are concretely achieved at this level. Trust is grounded in the use of certified XAI tools that ensure the quality and fidelity of the original explanation. These algorithmic explainability methods from Level 1 provide the foundation upon which all subsequent interpretations are built. LLMs serve exclusively as interpreters that make the original explanation accessible to less expert users—they do not modify or alter the explanation content, but rather translate it into comprehensible natural language. The ethical implications and contextual considerations are not generated by the LLM but emerge from human interaction when users ask specific questions about particular aspects of the concrete example being explained. This human-driven dialogue allows domain experts and stakeholders to explore ethical dimensions, fairness concerns, and societal impacts through targeted inquiries, ensuring that ethical considerations are grounded in human expertise rather than algorithmic assumptions.}

{In order to achieve these purposes, the role of LLMs is particularly useful, since they are increasingly becoming central in the field of XAI~\cite{bilal2025llms}, helping bridge the gap between complex model behavior and human understanding. Their integration into interactive conversational systems facilitates richer communication between users and AI systems by allowing natural language interfaces and interactive dialogue systems. However, including the use to directly explain machine learning models and the combination of explanations produced by other XAI methods, which are later refined using LLMs, is currently a significant challenge~\cite{zytek2024llms,bilal2025llms}.} XAI methods can harness the emergent capabilities of LLMs to facilitate more human-like communication. This is achievable thanks to LLMs' extensive background knowledge across various domains, enabling the generation of more intuitive explanations. In this context, technical explanations generated by other XAI methods can be transformed into more accessible, human-friendly narratives:

\begin{itemize}
    \item \textbf{Conversational interfaces for explanation:} LLMs facilitate dynamic dialogue-based interactions that go beyond static or one-shot explanations. Systems such as \textit{TalkToModel}~\cite{slack2023explaining} exemplify this paradigm. These systems allow users to iteratively query machine learning models using natural language, receive tailored explanations, and refine their understanding through follow-up questions. Compared to traditional explanation techniques, LLM-powered dialogue systems are inherently more user-centric and adaptable to diverse knowledge levels and tasks. Such interfaces emulate human-to-human explanation patterns, supporting the idea that explanations are inherently social, contrastive, and selective~\cite{bertrand2023selective}. Users receive justifications and, using LLM's capabilities, can probe, challenge, and request clarification, actions crucial to trustworthy machine learning systems.
    \item \textbf{Dimensions of interactivity in LLM-based XAI:} Building on the taxonomy proposed in~\cite{bertrand2023selective}, LLM-driven systems typically operate across three major axes of interactivity:
        \begin{enumerate}
            \item \textbf{Selective explanations:} Users can request specific explanations (why did it classify this email as spam?) or explore alternative reasoning (why not label it as promotion?), dynamically tailoring the content.
            \item \textbf{Mutable explanations:} Users can interactively explore model sensitivity and decision boundaries by simulating counterfactuals or adjusting input.
            \item \textbf{Dialogic explanations:} These interactions mimic human conversation, with iterative exchanges that refine both the user's and model's understanding of the task context.
        \end{enumerate}
\end{itemize}

\section{Case Studies: Applying the Multilevel Framework with LLMs}
\label{sec:examples}

\begin{figure}
    \centering
    \includegraphics[width=\textwidth]{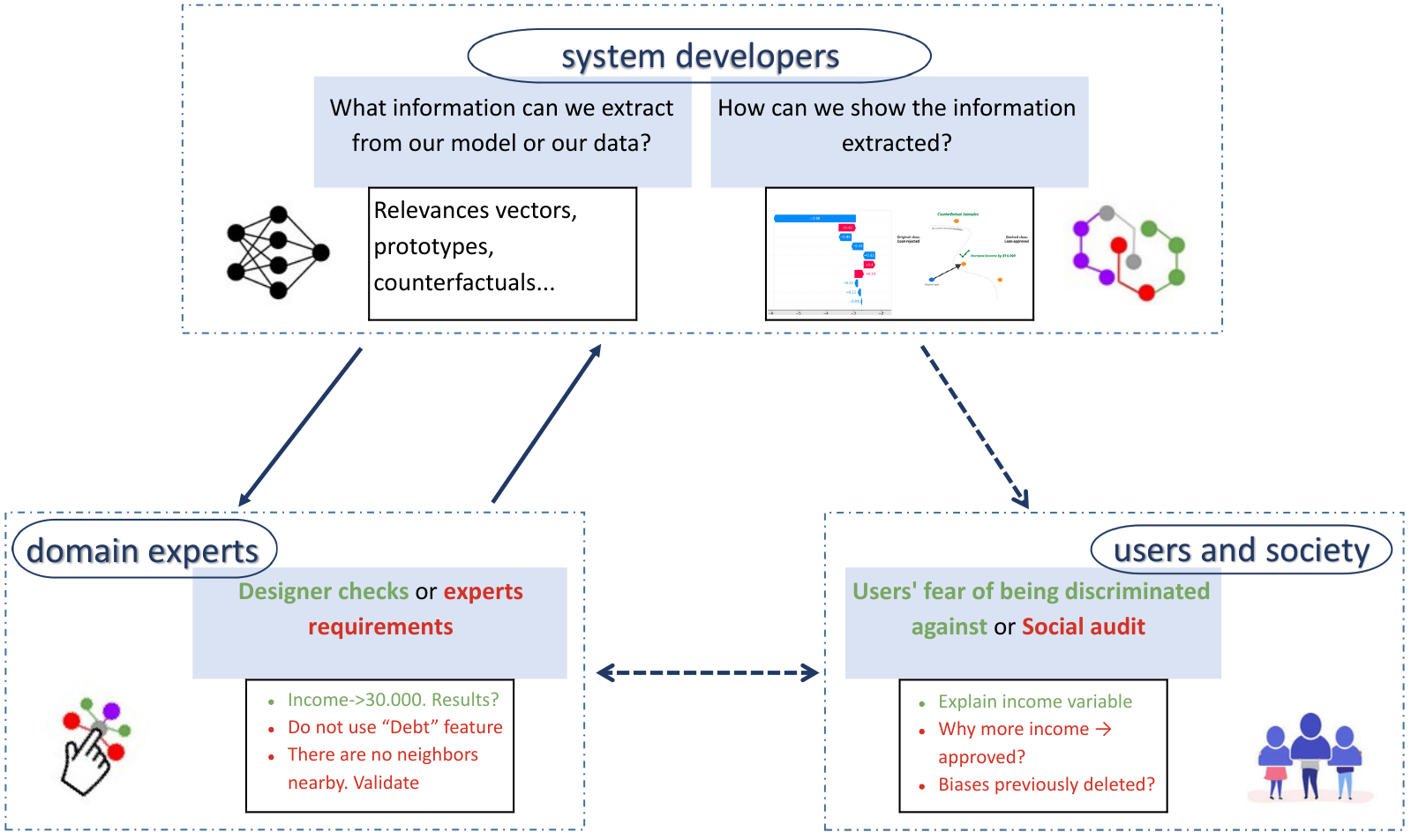}
    \caption{General overview of our framework for the case studies of approving a loan.}
    \label{fig:GeneralCaseStudies}
\end{figure}

To illustrate the application of our multilevel explainability framework in practice, we now present a set of concrete case studies. Each example is drawn from a common scenario—loan approval—to show how explanations can be constructed across different levels and enriched using Large Language Models (LLMs). These cases show how LLMs can support the generation of explanations by translating technical output into user-friendly narratives, facilitating interactive dialogue, and contextualizing decisions for greater societal understanding. \textcolor{black}{We also have designed Figure~\ref{fig:GeneralCaseStudies}, which implements our methodology for the specific case of granting a loan, from which we derive the three use cases implemented.}

The first case study shows how LLMs can generate explanations grounded in model outputs and domain knowledge, helping users interpret black-box predictions through structured, informative narratives. The second highlights the role of LLMs in enabling interactive, dialogue-based explanations, even when the underlying model is inherently interpretable, by adapting responses to user questions in natural language. The third illustrates how LLMs can enhance prototype- and counterfactual-based reasoning, providing intuitive and actionable insights into model behavior.

Together, these examples show how LLMs can effectively support explanation generation at the algorithmic, interaction, and societal levels of our framework. The first case leverages SHAP values and LLMs for user-centered personalized explanations of a black-box model. The second integrates interpretable models, such as decision trees, with LLMs to provide clear, conversational justifications. The third uses prototypical and counterfactual reasoning, enriched with natural language explanations, to help users understand and engage with the boundaries of model decisions. Collectively, these scenarios showcase the potential of high-quality, LLM-enhanced explanations to foster transparency, trust, and accountability in AI systems.

The following examples are generated using OpenAI’s GPT-4.5 model~\cite{openai_learning_2024}. The model was prompted to generate explanations based on the provided context and user queries.
\textcolor{black}{It is important to note that for these case studies, the use of GPT-4.5 is purely illustrative, as the data employed is non-sensitive and publicly available. In real-world applications involving private or sensitive information, on-premise LLM implementations should be prioritized to ensure ethical compliance and privacy protection.}

The first case study exemplifies how LLMs can enhance explanations at the algorithmic and domain-based level by translating complex model outputs—such as SHAP values—into coherent, user-friendly narratives.
In this scenario, a machine learning model is used to assess loan applications and its predictions are interpreted using feature attribution techniques.
LLMs play a central role in converting these numerical explanations into natural language justifications that incorporate relevant domain knowledge, such as financial thresholds and applicant profiles.

This allows users, including those without technical expertise, to understand which factors influenced the decision and how adjustments to their input data might change the outcome.
By focusing the explanation on both model logic and contextual knowledge, this case highlights the ability of LLMs to act as effective mediators between technical transparency and meaningful communication. \textcolor{black}{In this context, we have also introduced the interaction of an expert who asks the machine to modify its behavior, as the use of a specific variable to bias is not appropriate according to company policy. In addition to interpreting the explanation without this variable, the system proposes changes that the human must perform to avoid this behavior in future interventions.}

The second case study focuses on an inherently interpretable model—a decision tree—which naturally offers rule-based explanations for its predictions. Unlike black-box models that require post-hoc interpretation techniques, decision trees provide transparent logic by design, outlining decision paths through a sequence of conditions based on input features. This case complements the first by showing how interpretable models can still benefit from integration with LLMs. Here, the decision logic is not only readily accessible but can be translated into human-readable, conversational formats, making it more digestible for a wider range of users. Through LLMs, even straightforward rule sets can be personalized, contextualized, and clarified in response to user queries. This supports the delivery of explanations at both the algorithmic and the interaction levels, reinforcing understanding, promoting user trust, and allowing clarification without requiring in-depth technical knowledge. Case study 2 integrates well with the previous SHAP and LLM examples, showing a rule-based alternative that is easy to explain.

\begin{tcolorbox}[title=Case study 1: LLM-Based Explanation in a Loan Approval Scenario.,  colback=blue!5!white, colframe=blue!50!black]
\textbf{Context:} A financial institution uses a machine learning model to evaluate loan applications. An applicant has been denied a loan and requests an explanation. The system internally uses SHAP values to interpret the model outputs. Credit score 580 (SHAP value: \texttt{-0.35}); Annual Income \$28{,}000 (SHAP value: \texttt{-0.25}); Existing Debt: \$10{,}000 (SHAP value: \texttt{-0.15}); Employment Length (SHAP value: \texttt{+0.10}).

\textbf{User:} Why was my loan denied?

\textbf{System:} \textit{Your loan was denied because key factors negatively impacted your application. The most influential was your credit score of 580, which contributed -0.35 to the decision. Your annual income of \$28,000 and existing debt of \$10,000 also reduced your approval likelihood, contributing -0.25 and -0.15 respectively. Your employment length had a small positive impact (+0.10), but it was not enough to offset the negatives. These values reflect how each factor influenced the model’s output based on similar past applications.}

\textbf{User:} What if I had a higher income?

\textbf{System:} \textit{If your income were higher, the SHAP value for that feature would likely be less negative or even positive, which could improve your overall evaluation. For example, if your income increased to around \$50,000, the contribution might shift closer to 0 or above, reducing the model's confidence in denying the loan. However, the final outcome would still depend on how other factors, like credit score and debt, interact with the higher income. Improving multiple factors generally has a greater effect than changing one alone. We can run a simulation if you'd like to see the impact of a specific income level.}

\textbf{Expert: } Our policy prevents us from discriminating based on income, so the system should not consider this factor in the decision. What do you think then?

\textbf{System:} \textit{Thank you for pointing this out. If policy prohibits using income as a factor, then the model should not include annual income in its decision-making process. In that case, the main contributors would be your credit score (-0.35), existing debt (-0.15), and employment length (+0.10). The denial would still be primarily due to the low credit score and current debt burden. To remain compliant, the system needs to be updated so that income is excluded from both training and evaluation, ensuring that only permitted features influence loan decisions.}

\end{tcolorbox}

\begin{tcolorbox}[title=Case study 2: Interpretable Model Using a Decision Tree., colback=green!5!white, colframe=green!50!black]
\textbf{Context:} A decision tree is used to determine loan approval based on a few key features: credit score, annual income, and length of employment. The following is a simplified version of such a tree.

\begin{center}
\begin{tikzpicture}[level distance=1.5cm,
  level 1/.style={sibling distance=6cm},
  level 2/.style={sibling distance=4cm},
  level 3/.style={sibling distance=3cm}]
\node {Credit Score $\geq$ 600?}
  child {node {Income $\geq$ \$40{,}000?}
    child {node {Approved}
      edge from parent node[left] {Yes}}
    child {node {Employment $\geq$ 2 years?}
      child {node {Approved}
        edge from parent node[left] {Yes}}
      child {node {Denied}
        edge from parent node[right] {No}}
      edge from parent node[right] {No}}
    edge from parent node[left] {Yes}}
  child {node {Denied}
    edge from parent node[right] {No}};
\end{tikzpicture}
\end{center}

\textbf{Interpretation:} If an applicant has a credit score of at least 600, their income and employment length are further evaluated. An income above \$40{,}000 leads to automatic approval. If income is lower, then employment history is checked—if it's at least 2 years, the loan is still approved. Otherwise, it's denied. Any applicant with a credit score below 600 is automatically denied.

\textbf{Applicant Example:} The aplicant has the following features: Credit Score of 580, Income of \$28{,}000 and Employment of 6 years.

\textbf{Decision:} \texttt{Denied} — because the credit score is below the minimum threshold required for consideration.

\textbf{User:} Why was my loan denied?

\textbf{System:} \textit{Your loan was denied because your credit score is 580, which is below the minimum threshold of 600 required for further evaluation. According to our decision process, applicants must have a credit score of at least 600 to be considered for approval. Since this condition was not met, the system did not proceed to evaluate your income or employment length. This is an automatic decision based on the structure of the approval criteria.}

\textbf{User:} How can I improve my chances in the future?

\textbf{System:} \textit{To improve your chances, focus on raising your credit score above 600, as this is the first requirement for loan consideration. You can do this by paying bills on time, reducing outstanding debt, and avoiding new credit inquiries. Maintaining a stable income and long-term employment will also strengthen your application once the credit score threshold is met. Regularly monitoring your credit report can help you track progress and correct any errors. Once your score improves, you’ll be eligible for full evaluation under our approval criteria.}

\end{tcolorbox}

\begin{tcolorbox}[title=Case study 3: Prototypical and counterfactual explanations., colback=orange!10!white, colframe=orange!70!black]
\textbf{Context:} A machine learning model is used to evaluate loan applications. The model is trained as an autoencoder to learn a representation of the data, and then with the training data, a new sample is predicted as either \texttt{Approved} or \texttt{Denied} depending on the nearest neighbors in the learned representation. The model uses features such as credit score, annual income, and duration of employment.

\textbf{Interpretation:} The model generates two samples, one as a prototype of the predicted class and the other as a counterfactual example. The prototype represents the typical characteristics of the same class as the predicted sample, while the counterfactual example represents a hypothetical scenario where the sample would belong to the opposite class, all belonging to the trained model.

\textbf{Applicant Example:}  
\begin{itemize}
  \item Credit Score: 580  
  \item Income: \$28{,}000  
  \item Employment: 6 years  
\end{itemize}

\textbf{Decision:} \texttt{Denied} — because the closes sample in the learned representation is a prototype of the \texttt{Denied} class: Credit Score: 550, Income: \$25{,}000, Employment: 5 years. The counterfactual example is a sample that would be classified as \texttt{Approved}: Credit Score: 600, Income: \$40{,}000, Employment: 2 years.  

\textbf{User:} Why was my loan denied?

\textbf{System:} \textit{Your loan was denied because the system found that your application is more similar to past applicants who were also denied. For example, your profile (Credit Score: 580, Income: \$28,000) closely matches a typical denied case with a lower credit score and income. The model also created a “what-if” example showing that with a credit score of 600 and an income of \$40,000—even with less job experience—you would likely have been approved. This helps us understand what changes might lead to a different outcome.}

\end{tcolorbox}

The third case study explores how prototype-based and counterfactual explanations can be used to provide intuitive insights into model behavior, particularly when the underlying model operates in a latent feature space. In this scenario, a machine learning model trained as an autoencoder evaluates loan applications by comparing new instances to previously learned patterns. Instead of explaining individual feature contributions, the model identifies a typical (prototype) case similar to the applicant and a contrasting (counterfactual) example that would have resulted in a different decision. This dual perspective helps users understand why a decision was made and what changes could lead to an alternative outcome. LLMs play a critical role in translating these technical comparisons into accessible, human-centered narratives. This supports users in understanding both the model's rationale and actionable pathways, bridging levels 1 and 2 of the framework. Furthermore, by articulating these scenarios in natural language, LLMs help frame the explanation in terms relevant to fairness, opportunity, and future planning, core aspects of societal transparency at level 3.

To assess the practical applicability of our multilevel framework, we analyze how each of the three case studies maps to the different levels of explainability. This structured analysis allows us to assess the role and contribution of each level, which is algorithmic, human-centered, and societal, in generating trustworthy and audience-adapted explanations. By aligning specific elements of the case studies with the corresponding levels, we show how the framework supports layered, progressive understanding, and how LLMs play a unifying role across the spectrum of explanation needs.

{The three levels of our framework can be distinguished on each of the three example as follows:}
\begin{enumerate}
    \item {\textbf{Level 1 (Algorithmic and Domain-Based Explainability):} In the first case, the SHAP values provide a model-based explanation of how each feature contributes to the loan decision. In the second case, the decision tree serves as an inherently interpretable model that explains the decision-making process through clear rules. In the third case, the system provides a prototype and a counterfactual sample that illustrate the learned representation of the data.}
    \item {\textbf{Level 2 (Human-Centered Explainability):} This level is exemplified in the first case with the interpretation of the SHAP values, where an end-user can interpret these values in the context of their loan application. In the second case, the decision tree provides a straightforward explanation that can be easily understood by an end-user with knowledge of how decision trees work. The third case illustrates how the prototype and counterfactual examples can be interpreted by the user to understand the model's decision-making process by comparing their own case to the prototype and counterfactual instances.}
    \item {\textbf{Level 3 (Social Explainability):} This level is exemplified in the three cases by the LLM-based system's ability to engage in dialogue with the user, providing clearer explanations and allowing for follow-up questions. This system gets as input the explanations generated in the first level and generates an interactive explanation that is tailored to the user's needs in natural language.}
\end{enumerate}

{The ability of LLMs to articulate ethical concerns, simulate diverse perspectives, and produce policy-aware narratives opens new avenues to align AI behavior with regulatory and civic expectations. In this regard, social explanations can serve as communication tools while providing mechanisms for accountability, traceability, and compliance. Potential use cases include integration with algorithmic audit platforms, explainability requirements in AI Act-like legislation, and participatory mechanisms such as citizen panels or public hearings. By explicitly recognizing these applications, this level situates explainability within the broader landscape of responsible AI, where technical transparency and social legitimacy must go hand in hand.}

To complement the descriptive analysis of each case study, Table~\ref{tab:case-level-mapping} provides a visual summary of how the proposed multilevel explainability framework applies in the three scenarios. Each case integrates elements from all three levels: algorithmic and domain-based explainability (Level 1), human-centered interaction (Level 2), and social explainability through LLM-mediated communication (Level 3). This mapping reinforces the generalizability and modularity of the framework, showing how different technical approaches and explanation strategies can be coordinated to address diverse stakeholder needs and build trust between user contexts~\cite{ali2023explainable,herrera2025reflections}.

\begin{table}[h]
\centering
\caption{Mapping of Case Studies to Explainability Levels.}
\label{tab:case-level-mapping}
{\small
\begin{tabular}{|p{4cm}|p{2.2cm}|p{2.5cm}|p{2.2cm}|}

\hline
\textbf{Case Study} & \textbf{Level 1: Algorithmic} & \textbf{Level 2: Human-Centered} & \textbf{Level 3: Social} \\
\hline
\textbf{Case 1: SHAP + LLM} & \checkmark & \checkmark & \checkmark \\
\hline
\textbf{Case 2: Decision Tree + LLM} & \checkmark & \checkmark & \checkmark \\
\hline
\textbf{Case 3: Prototype/Counterfactual + LLM} & \checkmark & \checkmark & \checkmark \\
\hline
\end{tabular}
}%
\end{table}

\section{Conclusions}
\label{sec:5}

{In this work, we introduce a multilevel framework for XAI that addresses the needs of domain experts and broader societal audiences. The first level focuses on model-based explanations, combining inherently interpretable models and post hoc analyses to reveal how input drives outputs and how domain knowledge is incorporated. The second level supports human-centered explanations, allowing stakeholders to interact with, query, and adapt the explanation content through dialogue. The third level delivers socially grounded narratives that foster public trust, regulatory compliance, and collective accountability. By structuring explanations in this way, our framework ensures that each successive level builds on the fidelity and technical rigor of the previous level while extending accessibility to non-technical users.}

{A central contribution of our approach is the integration of LLMs as mediators between technical outputs and user-oriented explanations. In our case study of a loan-denial scenario, we show how SHAP values, tree-based models, or example-based explanations can be transformed by an LLM into a conversational narrative that highlights the most influential features, contextualizes them within legal and ethical considerations, and addresses user-specific concerns. This integration preserves the true behavior of the underlying mode while generating clear, personalized explanations that empower end-users and domain experts alike.}

{Beyond individual comprehension, fostering societal trust in AI requires that systems explain why decisions are made and how these decisions align with shared values and legal expectations. Our framework treats XAI as a societal instrument: interpretable models offer auditable logic, post-hoc techniques expose complex relationships, and LLM-based interfaces bridge the gap to public understanding. Achieving this vision allows interdisciplinary collaboration to identify and incorporate diverse ethical norms and to design participatory mechanisms that capture community perspectives and ensure that explanations remain meaningful to citizens.}

While the proposed multilevel framework offers a structured approach to audience-centered explainability, it remains a conceptual foundation that requires further empirical validation. The illustrative case studies provide plausibility but are not exhaustive across domains or user types. Additionally, the use of LLMs introduces challenges related to factual consistency, bias amplification, and explainability of the LLMs themselves.

{Looking ahead, several research directions remain open. First, there is a critical need for standardized qualitative and quantitative metrics to evaluate the effectiveness, usability, and ethical adequacy of explanations at all three levels. Second, adaptive systems could dynamically tailor explanation types to individual user profiles, learning contexts, or decision criticality. Third, operationalizing social explanations on a scale will require integrating participatory audit systems, regulatory dashboards, and community feedback loops. Finally, since LLMs play an increasingly central role in these explanation pipelines, ensuring their fairness, robustness, and alignment with societal values is essential. Addressing these challenges will help transform our three-level framework from a conceptual roadmap into a practical, trustworthy solution for real-world AI systems.}

{An important avenue for future research lies in deepening the methodological rigor of Level 1 within the framework, particularly in connection with RED XAI principles. While current case studies show how algorithmic transparency can clarify model behavior using LLMs, future work could formalize how such explanations can be used for understanding and also for iterative model improvement. This would involve creating structured feedback loops between the explanation process and model design, enabling researchers and developers to diagnose, refine, and validate model behavior during training and deployment. Using explainability as a tool for debugging and enhancement, rather than merely post hoc justification, this approach can further advance the reliability and robustness of AI systems.}

As AI systems become increasingly embedded in critical decision-making domains, the future of explainability must be grounded not only in technical advancements but also in real-world deployment constraints and stakeholder diversity. Future research should explore how multilevel explanations can be integrated into operational AI pipelines in domains such as healthcare, finance, and public services, where transparency and accountability are paramount. This includes designing user studies across different stakeholder groups to assess how explanations influence understanding, trust, and behavior. In addition, interdisciplinary collaboration will be essential, with legal scholars, social scientists, and domain experts, to ensure that the explanations are not only comprehensible, but also normatively sound. Ultimately, developing frameworks that link technical fidelity with legal robustness and civic legitimacy will be critical for embedding AI systems within sustainable and trustworthy governance structures.

\section*{Declarations}

\subsection*{Ethics approval and consent to participate}
Not applicable.

\subsection*{Consent for publication}
Informed consent for publication was obtained from all participants.

\subsection*{Availability of data and materials}
Not applicable.

\subsection*{Competing interests}
The authors declare no competing interests.

\subsection*{Funding}

This research has been developed within the R\&D project CONFIA (PID2021-122916NB-I00), funded by MICIU/AEI/10.13039/501100011033/ and FEDER, EU, and the ICT-NETWORK collaboration program (CU2019N ET033A105) of the Council of Flemish Universities of Belgium (VLIR-UOS). In addition, it is part of the project ``Ethical, Responsible, and General Purpose Artificial Intelligence: Applications in Risk Scenarios'' (IAFER) Exp.:TSI-100927-2023-1, funded through the creation of university-industry research programs (Enia Programs), aimed at the research and development of artificial intelligence, for its dissemination and education within the framework of the Recovery, Transformation, and Resilience Plan of the European Union Next Generation through the Ministry of Digital Transformation and the Civil Service. Also, it is partially supported by Knowledge Generation Projects, funded by the Spanish Ministry of Science, Innovation, and Universities of Spain under the project PID2023-150070NB-I00. As well as, the project ``Theoretical contributions to AI in handling complex data problems'' under the National Program of Science, Technology, and Innovation in Automation, Robotics, and Artificial Intelligence, CUBA.

\subsection*{Authors' contributions}
Marilyn Bello and Rafael Bello conducted the initial work and conceptualized the study.  
Iván Sevillano‑García and Francisco Herrera developed and carried out the experiments.  
All authors participated in writing the manuscript and critically reviewed and approved the final version for publication.

\subsection*{Acknowledgements}

Not applicable.

\subsection*{Declaration of AI-assisted technologies in the writing process}

During the preparation of this work, the authors used large language models to improve the readability and language of the manuscript. After using this tool/service, the authors reviewed and edited the content as needed and assumed full responsibility for the content of the published article.

\bibliography{bibliography}

\end{document}